# A Feature Learning and Object Recognition Framework for Underwater Fish Images

Meng-Che Chuang, Jenq-Neng Hwang, and Kresimir Williams

*Abstract*—Live fish recognition is one of the most crucial elements of fisheries survey applications where vast amount of data are rapidly acquired. Different from general scenarios, challenges to underwater image recognition are posted by poor image quality, uncontrolled objects and environment, as well as difficulty in acquiring representative samples. Also, most existing feature extraction techniques are hindered from automation due to involving human supervision. Toward this end, we propose an underwater fish recognition framework that consists of a fully unsupervised feature learning technique and an error-resilient classifier. Object parts are initialized based on saliency and relaxation labeling to match object parts correctly. A non-rigid part model is then learned based on fitness, separation and discrimination criteria. For the classifier, an unsupervised clustering approach generates a binary class hierarchy, where each node is a classifier. To exploit information from ambiguous images, the notion of partial classification is introduced to assign coarse labels by optimizing the "benefit" of indecision made by the classifier. Experiments show that the proposed framework achieves high accuracy on both public and self-collected underwater fish images with high uncertainty and class imbalance.

## I. Introduction

IMAGE PROCESSING and analysis techniques for underwater cameras have drawn increasing attention since they enable a non-extractive and non-lethal approach to fisheries survey [1-6]. For instance, by using a combination of cameras and mid-water trawl, known as the Cam-trawl [7], fish schools are sampled by capturing images or videos while they pass through the trawl. The camera-based sampling approach not only conserves depleted fish stocks but also provides an effective way to sample a greater diversity of marine animals. This approach, however, generates vast amounts of data very rapidly. An automatic image processing system is thus critically required to make such a sampling approach practical.

Manuscript received March 12, 2015, accepted February 21, 2016. This work was supported by the National Marine Fisheries Services' Advanced Sampling Technology Working Group, National Oceanic and Atmospheric Administration, Seattle, WA, USA.

M.-C. Chuang and J.-N. Hwang are with the Department of Electrical Engineering, University of Washington, Seattle, WA 98195 USA (e-mail: mengche, hwang@uw.edu).

K. Williams is with the Alaska Fisheries Science Center, National Oceanic and Atmospheric Administration (NOAA), Seattle, WA 98115 USA (e-mail: kresimir.williams@noaa.gov).

Toward this end, we have developed techniques that analyze the collected data by automatic object segmentation, size estimation, counting and tracking [8-10]. Based on this, an automatic camera-based fisheries survey system can be realized by developing a reliable species identification algorithm that allows for monitoring the species composite and assessing fish stocks as well as the ecosystem.

While object recognition in various contexts has been well investigated in image processing and computer vision communities, there exist fundamental challenges to identifying live fish in an unconstrained natural habitat. Like most underwater imagery scenarios, one challenge is posted by the low image quality caused by fast attenuation of light in the water, poor control over illumination, the ubiquitous organic debris, etc. While capturing images for freely-swimming fish, there is a high uncertainty in many of the data due to low image quality, non-lateral fish views or curved body shapes. This seriously degrades the recognition performance since some critical information may be lost. Even without uncertainty, fish share a strong visual correlation among species. Common image features for object recognition are usually not sufficiently discriminative in this case.

Another common challenge in applications of statistics or machine learning techniques is the existence of uncertain or missing samples. One strategy to handle this fact is partial classification, i.e., allowing indecision made by the classifier in certain regions in the data space. Partial classification has shown its effectiveness in various practical applications [11-13]. However, the importance of rejected instances is gone since no information about the data is retrieved. Besides, there are yet no systematic methods proposed to determine the criteria of decision making. Since objects can be naturally categorized into higher groupings of classes based on either domain knowledge or visual similarity, the recognition algorithm would be favorable by obtaining a hierarchical relation among classes automatically and then providing a coarse-to-fine categorization that can retrieve partial information from those uncertain data.

In this paper, we propose a novel feature learning and object recognition framework that addresses the challenges described above, as shown in Fig. 1. One advantage of the proposed framework is that it uses fully unsupervised algorithms to learn the features and class correlation, and thus provides an automatic solution for practical recognition systems. Specifically, the contributions of this paper include: 1) a novel non-rigid part model that represents both appearance and geometric attributes of the fish body; 2) an unsupervised learning algorithm of non-rigid part model based on systematic part initialization and an EM-like alternating optimization algorithm; 3) a novel hierarchical partial classification that successfully handles data uncertainty and class imbalance; 4) a formal approach that determines the decision criteria based on an optimization formulation.

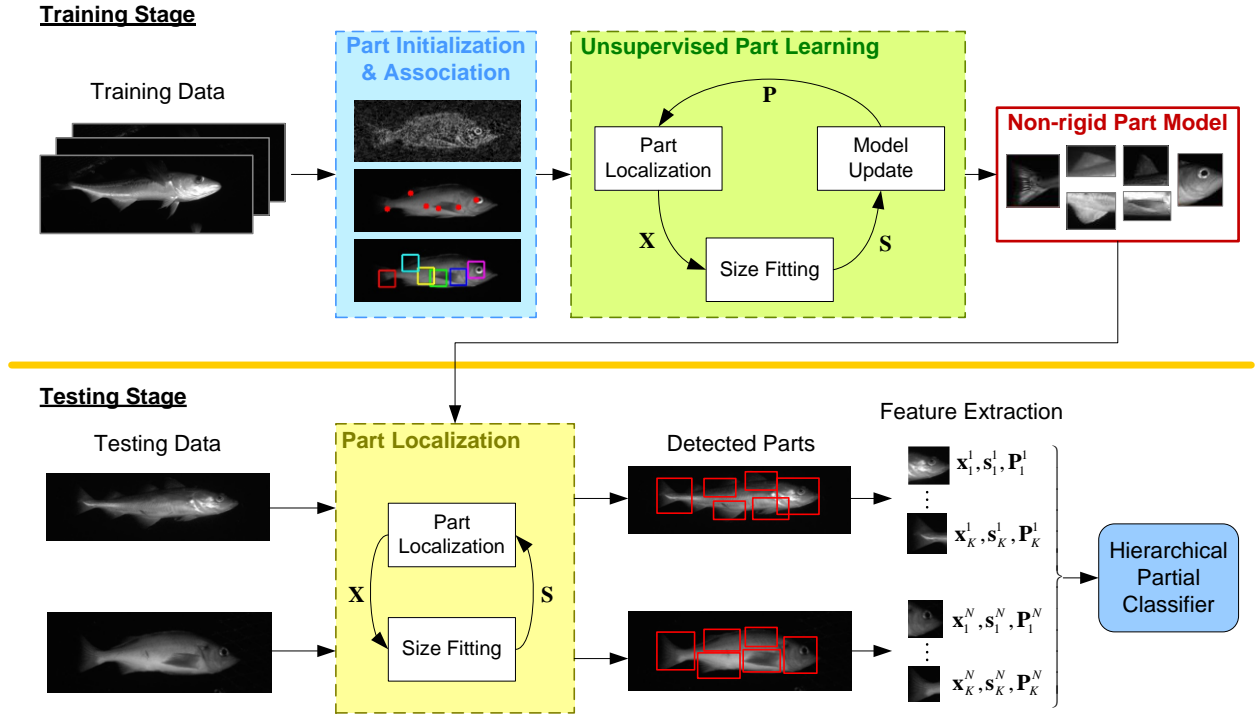

Fig. 1. Overview of the proposed unsupervised non-rigid part learning algorithm. In the training stage, object parts are initialized and associated across training image. The non-rigid part model is then learned from images via an unsupervised algorithm. In the testing stage, the model locates informative object parts in each testing image. The location, size and appearance of each part are extracted as features. Finally, these features are used to train a fine-grained object classifier.

The rest of this paper is organized as follows. Section II gives a brief review of the related work. Section III describes the problem formulation. Section IV introduces the unsupervised non-rigid part model learning algorithm. Section V describes the hierarchical partial classification method. Section VI reports the experimental results on fish species recognition, and the conclusion is given in Section VII.

## II. RELATED WORK

### A. Fish Recognition

Live fish recognition is one of the most crucial elements in camera-based fisheries survey systems [2-5]. Similar to most recognition frameworks, the successful extraction of informative features is the key to enhancing fish recognition performance. Existing feature extraction techniques are divided into two categories, namely the supervised and unsupervised methods. Supervised methods represent a fish by pre-specified features that adopt common low-level image descriptors such as contour shape. For instance, Lee et al. [5] used a curvature analysis approach to locate critical landmark points. The contour segments of interest were extracted based on these landmark points to achieve satisfactory shape-based species classification results. In their subsequent work [3], the features were further extended to include several shape descriptors such as Fourier descriptors, polygon approximation and line segments. A power cepstrum technique was developed in order to improve the categorization speed using contours represented in tangent space with normalized length. Spampinato et al. [2] proposed fish descriptors that consider appearance attributes such as texture in addition to contour shape. With the huge diversity of fish, however, one set of features designed for some species is not guaranteed to be discriminative for other species. Moreover, manually selected features may lead to suboptimal recognition performance even based on domain knowledge.

On the other hand, unsupervised methods learn informative features directly from the images. Some approaches in this category can be found in the literature of fine-grained object recognition [14-18] or discriminative mid-level image patch discovery for scene recognition [19, 20]. There are also other branches of unsupervised methods based on, for example, traditional feature selection theories [21]. We have conducted a systematic comparison between supervised and unsupervised approaches and shown that the unsupervised approaches in general lead to better recognition performance [22]. Based on this, we extend the unsupervised feature learning algorithm in this paper by proposing a novel non-rigid part model, which considers part geometry and can be learned in a fully unsupervised manner.

### B. Classification with Data Uncertainty

Traditional strategies in statistics for handling uncertain data include discarding these samples or performing imputation, where estimates are used to fill in the missing values. Some work integrated the classification formulation with the concept of robust statistics by assuming different noise distributions [12, 13]. Huang et al. [4] used a hierarchical classifier constructed by heuristics to control the error accumulation. However, errors still propagated to the leaf layer once they occurred. For partial classification, Ali et al. [12] proposed to determine whether to make a decision based on data mining techniques. An evidential classification approach [23, 24] was presented to commit incomplete objects that are hard to classify to the associate set of classes with belief functions. Baram [13] introduced a benefit function for evaluating the deferred decisions and search for the optimal decision criterion exhaustively. We generalized the definition of benefit function, and propose a novel formulation based on exponential functions that systematically helps select the decision criteria for a partial classifier [25].

*C. Comparison to Previous Work*

This paper is extended from our previous work on unsupervised feature extraction [22] and hierarchical partial classification [25] for fish recognition. The main differences of the method described in this paper from the previous work can be found in two aspects. One is the use of saliency operation to initialize part locations instead of arbitrarily splitting the bounding box. This systematically provides a reasonable starting position for each part that can avoid local optima during the alternating optimization. The other aspect is the part alignment by relaxation labeling process. Based on some topological constraints, parts are successfully matched from one image to another despite the pose variations. This step not only ensures the correctness during part feature learning but also offers a higher spatial flexibility comparing to existing template-based methods [14].

### III. NON-RIGID PART MODEL

Given a set of training images $\mathbf{I} = \{I^1,...,I^N\}$, the goal is to discover discriminative features for the objects in terms of their subordinate categories. Let $M = \{\mathbf{P}, \mathbf{X}, \mathbf{S}\}$ denote a model that consists of part appearance $\mathbf{P} = \{\mathbf{P}_1,...,\mathbf{P}_K\}$, part center locations $\mathbf{X} = \{\mathbf{X}^1,...,\mathbf{X}^N\}$ and part sizes $\mathbf{S} = \{\mathbf{S}^1,...,\mathbf{S}^N\}$, where $K$ is the number of object parts. For each image $I^m$ we denote $\mathbf{X}^m = \{\mathbf{x}_1^m,...,\mathbf{x}_K^m\}$ and $\mathbf{S}^m = \{\mathbf{s}_1^m,...,\mathbf{s}_K^m\}$ accordingly. The location and size of each part is normalized with respect to the image size, i.e., $x_i, s_i \in [0,1] \times [0,1]$. The model $M$ is referred to as a *non-rigid part model* since it describes common parts and allows for deformation in both position and scale. Based on this, the problem of learning such a model can be written as a constrained minimization programming problem:

$$(\mathbf{P}^*, \mathbf{X}^*, \mathbf{S}^*) = \arg\min J(\mathbf{P}, \mathbf{X}, \mathbf{S}) \quad (1)$$

$$\text{s.t.} \quad 0 \leq \mathbf{s}_i^m \leq 1, \quad i=1,...,K, \quad m=1,...,N \quad (2)$$

$$0 \leq \mathbf{x}_i^m \pm (1/2)\mathbf{s}_i^m \leq 1, \quad i=1,...,K, \quad m=1,...,N, \quad (3)$$

where $J(\mathbf{P}, \mathbf{X}, \mathbf{S})$ is the objective function of the model. Note that (2) and (3) denote entry-wise inequalities for $\mathbf{x}_i^m$ and $\mathbf{s}_i^m$.

To model both the appearance and geometry of object parts, the objective function $J(\mathbf{P}, \mathbf{X}, \mathbf{S})$ takes three factors into account: 1) *fitness*, which computes the appearance similarity between the model and an image region; 2) *separation*, which guides the detected parts to match as disjoint as possible regions instead of concentrating on a few regions of the image; 3) *discrimination*, which encourages the selected parts to have distinct appearance from each other in order to capture as many aspects of the object appearance.

*A. Fitness*

Image regions corresponding to a certain object part have high appearance similarity with the model. The fitness cost is thus calculated by the distance between the part appearance $\mathbf{P}_i$ and the appearance of a rectangular region in image $I^m$ defined by the center location $\mathbf{x}_i^m$ and size $\mathbf{s}_i^m$. We denote this region by $I_i^m$, then the fitness cost is given by,

$$J_{fitness} = \sum_{m=1}^{N} \sum_{i=1}^{K} d(\mathbf{P}_i, \phi(I(\mathbf{x}_i^m, \mathbf{s}_i^m))), \quad (4)$$

where $\phi(\cdot)$ denotes the feature descriptor for an image region, and $d(\mathbf{P}, \mathbf{Q})$ is the distance between two appearance feature vectors $\mathbf{P}$ and $\mathbf{Q}$. For the following part of this paper, we denote $I(\mathbf{x}_i^m, \mathbf{s}_i^m)$ by $I_i^m$ for convenience.

*B. Separation*

The separation cost enforces the parts to cover the maximum area of the whole object. This is achieved by minimizing the total overlapping rate of the image regions defined by the location-size tuples $(\mathbf{x}_i^m, \mathbf{s}_i^m)$ and $(\mathbf{x}_j^m, \mathbf{s}_j^m)$ for $i \neq j$.

$$J_{separation} = \sum_{m=1}^{N} \sum_{i=1}^{K} \sum_{j \neq i} v(I_i^m, I_j^m), \quad (5)$$

The overlapping rate is defined as the area ratio between the intersection and union of two rectangles, i.e.,

$$v(I_i^m, I_j^m) = \frac{|I_i^m \cap I_j^m|}{|I_i^m \cup I_j^m|}. \quad (6)$$

We denote $v(I_i^m, I_j^m)$ by $v_{i,j}^m$ for the following part of this paper for convenience.

*C. Discrimination*

It is desired that the non-rigid part model covers every representative part of the object. As a result, the discrimination cost is introduced to encourage the maximization of the distance between each pair of part features $\mathbf{P}_i$ and $\mathbf{P}_j$, i.e.,

$$J_{discrimination} = -\sum_{i=1}^{K}\sum_{j=1}^{K} d(\mathbf{P}_i, \mathbf{P}_j), \quad (7)$$

where $d$ is the same distance metric as the one in (4).

*D. Objective Function*

Having the above cost functions, the final objective function is written as

$$J(\mathbf{P}, \mathbf{X}, \mathbf{S}) = J_{fitness} + J_{separation} + J_{discrimination}, \quad (8)$$

The non-rigid part model, i.e., part features, locations and sizes are trained by minimizing (8) over the given training set $\mathbf{I}$ using the proposed unsupervised learning algorithm described in the following section.

## IV. UNSUPERVISED FEATURE LEARNING

Now we have a non-rigid part model – features, locations and size of each part – that represents the object local appearance and configuration, as well as an objective function (8) to be minimized to find the model. An EM-like algorithm, i.e., alternating optimization, enables an unsupervised approach to learning such a model from the training images. With a systematic initialization technique, no human annotation for parts is required for learning the final feature descriptors.

*A. Part Initialization*

The effectiveness of alternating optimization guarantees only the convergence to local optima. To ensure a good solution can be obtained, we propose a systematic approach to initialize the part model. Note that most details that distinguish fine-grained categories match those parts which are prominent to humans' perception, such as beak of a bird, pedal of a flower, or tail fin of a fish. Saliency operators works perfectly for this purpose.

There have been a variety of techniques investigated for estimating image saliency. For the efficiency in dealing with vast data amount, we adopt the phase Fourier transform (PFT) approach described in [26]. Given an image, we calculate its 2-D discrete Fourier transform, which can be expressed by the magnitude term $M(u,v)$ and phase term $\Phi(u,v)$, i.e.,

$$F(I(x,y)) = M(u,v)e^{j\Phi(u,v)}. \quad (9)$$

The saliency is obtained by taking the inverse Fourier transform of only the phase term, i.e.,

$$s(x,y) = G_\sigma(x,y) * F^{-1}(e^{j\Phi(u,v)}), \quad (10)$$

where $G_\sigma(x,y)$ is a 2-D Gaussian filter with standard deviation $\sigma$. Non-maximal suppression is applied to extract local maxima from the saliency map. Here we use the object segmentation mask produced by [8] to discard salient points in the background. Note that using the given segmentation does not make our learning method supervised, since the masks can be easily generated by existing techniques such as the GrabCut segmentation [27]. Finally, we choose the top $K$ local maxima locations, each of which serves as an initial part. Examples of part initialization based on PFT saliency are shown in Fig. 2.

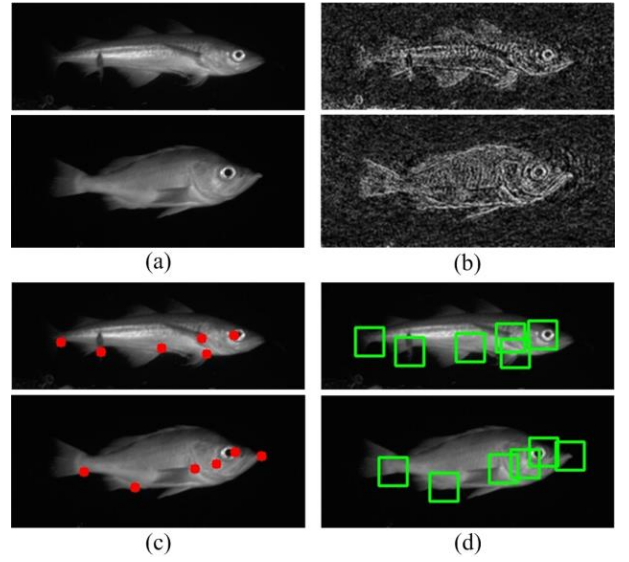

Fig. 2. Part initialization based on PFT saliency. (a) Input images. (b) PFT results of (a). (c) Points with top saliency values inside the segmentation masks. (d) Initial parts.

*B. Labeling-based Part Association*

Due to the pose variation, one object part may appear in different locations in two images. To ensure the correctness of learning, it is important to align the extracted points from one image to another. In the proposed method, we formulate part identification as a one-to-one association problem and apply the relaxation labeling process as follows.

Suppose a reference set contains $K$ part locations, denoted by $\mathbf{Y} = \{\mathbf{y}_1,...,\mathbf{y}_K\}$, and a candidate set also contains $K$ part locations $\mathbf{X} = \{\mathbf{x}_1,...,\mathbf{x}_K\}$. The goal of part identification is to find an optimal association from candidate parts to reference parts, which is similar to the matching problem between two sets of 2-D points that undergo some non-rigid deformation [28, 29]. The association can be expressed by a $(K+1) \times (K+1)$ binary matrix $\mathbf{\Pi}$, where $\pi_{ij} = 1$ when $\mathbf{x}_i$ is associated with $\mathbf{y}_j$ and $\pi_{ij} = 0$ otherwise. Each row of $\mathbf{\Pi}$ corresponds to a candidate part, and each column corresponds to a reference part. The augmented row and column denote the "outliers," which represent the case of no match for a certain candidate or reference part. Introducing the outlier notion brings in two advantages. Firstly, it allows for handling substantial pose variations or partial occlusions. Moreover, it facilitates the imposition of one-to-one match constraint between two part location sets.

In relaxation labeling, the binary constraint $\pi_{ij} \in \{0,1\}$ is relaxed to $\pi_{ij} \in [0,1]$. It has been proved that $\pi_{ij}$ converges to either 0 or 1 [29]. During the iterations, each entry $\pi_{ij}$ is updated by exploiting the contextual information, which is represented by the compatibility coefficient $r(i,j,k,l) \in [0,1]$. A high value of $r(i,j,k,l)$ corresponds to high association likelihood between $(\mathbf{x}_i, \mathbf{y}_j)$ and $(\mathbf{x}_k, \mathbf{y}_l)$. When the part configurations are non-rigid, the compatibility coefficient considers only local neighbors of both $\mathbf{x}_i$ and $\mathbf{y}_j$ [28, 29]. Here we define that two parts $\mathbf{x}_i$ and $\mathbf{x}_k$ are neighbors of each other only if $\mathbf{x}_i$ is one of the $k$-nearest parts of $\mathbf{x}_k$ and vice versa. In the experiments we set $k=3$, which gives promising results. For each part $\mathbf{x}_i$ the indices of its neighbors are denoted by a set $N_i$.

We define a novel compatibility coefficient that imposes pairwise geometric constraints between two neighboring parts as follows. To handle pose variation among part sets, we first project all part coordinates to the basis formed by the object's two principal components. Denote the projected coordinates of candidate and reference parts by $\mathbf{u}_i$ and $\mathbf{v}_j$, respectively. The associations $(\mathbf{u}_i, \mathbf{v}_j)$ and $(\mathbf{u}_k, \mathbf{v}_l)$ are more compatible if their distance between $\mathbf{u}_i$ and $\mathbf{u}_k$ is similar to the distance between $\mathbf{v}_j$ and $\mathbf{v}_l$. Also, if the vector direction from $\mathbf{u}_i$ to $\mathbf{u}_k$ is similar to the one from $\mathbf{v}_j$ to $\mathbf{v}_l$, they are more compatible. The distance and angle disparity is thus defined as

$$a(i,j,k,l) = \|\mathbf{u}_i - \mathbf{u}_k\|^2, \quad (11)$$
$$b(i,j,k,l) = (\cos\theta_{ik} - \cos\theta_{jl})^2. \quad (12)$$

Note here only the first principal component coordinates are considered to handle the left-right flipping of objects. The final compatibility coefficient between $(\mathbf{u}_i, \mathbf{v}_j)$ and $(\mathbf{u}_k, \mathbf{v}_l)$ is defined as $r(i,j,k,l) = \exp(-(a(i,j,k,l) + b(i,j,k,l)))$. Based on this, the support function $q_{ij}$ in each iteration is given by

$$q_{ij} = \sum_{l \in N_j} \sum_{k \in N_i} \pi_{kl} r(i,j,k,l). \quad (13)$$

Each entry $\pi_{ij}$ is then updated by

$$\pi_{ij} \leftarrow \pi_{ij} q_{ij} \Big/ \sum_{k=1}^{K} \pi_{ik} q_{ik}. \quad (14)$$

Alternated row-column normalization is performed after each relaxation update. It has been shown that this normalization process ensures that each row and column of $\mathbf{\Pi}$ sum to one according to Sinkhorn's theorem [29]. The final association for each target part is chosen upon convergence by the maximum $\pi_{ij}$ with respect to $j$.

### C. Unsupervised Part Model Discovery

The non-rigid object part model is learned by solving the optimization problem (1)–(3), where the objective function $J(\mathbf{P}, \mathbf{X}, \mathbf{S})$ is given by (8). In order to effectively update the

---

**Algorithm 1.** Non-Rigid Part Model Learning
1. **input:** images $\mathbf{I} = \{I_1, ..., I_N\}$, maximum iteration *maxIter*
2. **output:** part features $\mathbf{P}$, locations $\mathbf{X}$, sizes $\mathbf{S}$
3. initialize $\mathbf{P}, \mathbf{X}, \mathbf{S}$ by the PFT saliency
4. associate parts with reference by relaxation labeling
5. **for** $t = 1$ to *maxIter* **do**
6.     update $\mathbf{X}$ for each image based on (15), (16)
7.     update $\mathbf{S}$ for each image based on (19)-(21)
8.     update $\mathbf{P}$ based on (23)
9.     **if** *converged* **then**
10.         break
11.     **end if**
12. **end for**

---

variables $\mathbf{P}, \mathbf{X}, \mathbf{S}$ without human assistance in the loop, we adopt the alternating optimization that is similar to the expectation-maximization (EM) algorithm. In the proposed unsupervised part discovery algorithm, each iteration consists of three steps. The algorithm first updates the locations $\mathbf{X}$ (part localization) with the remaining variables fixed, then updates the sizes $\mathbf{S}$ (part size fitting) with the remaining variables fixed, and finally updates the part features $\mathbf{P}$ (part model learning) with the remaining variables fixed. For each training image the part locations and sizes are initialized based on the saliency detection and relaxation labeling procedure described in Section IV.A and IV.B. The appearance for each part is initialized by the average value of the corresponding block over the training set. The learning procedure of non-rigid part model is summarized in Algorithm 1.

*1) Part Localization*

In this step, the part features $\mathbf{P}$ and sizes $\mathbf{S}$ are given. By updating $\mathbf{X}$, we localize the sub-region that corresponds to each part in each image. The discrimination cost term in (7) becomes a constant since $\mathbf{P}$ is fixed for now. Hence the optimization problem (1)–(3) becomes

$$\min_{\mathbf{X}} \sum_{m=1}^{N} \left( \sum_{i=1}^{K} d(\mathbf{P}_i, \phi(I_i^m)) + \sum_{i=1}^{K} \sum_{j \neq i} v_{i,j}^m \right) \quad (15)$$

$$\text{s.t.} \quad 0 \leq \mathbf{x}_i \pm (1/2)\mathbf{s}_i \leq 1, \quad i = 1, ..., K. \quad (16)$$

Here the mean-shift algorithm [30] is adopted to solve for $\mathbf{X}$ as follows. In the $m$-th image, we start from the initial location $\mathbf{x}_i^m(0)$ for the $i$-th object part. The typical mean-shift algorithm is a gradient ascending optimization procedure, so the negative of objective function (15) is maximized instead. The gradient estimate is iteratively calculated by using pixels within the located sub-region. Given the gradient estimate, the part location is updated by

$$\mathbf{x}_i^m(t+1) = \frac{\sum_{j=1}^{n_p} k(\mathbf{z}_j - \mathbf{x}_i^m(t)) w_j (\mathbf{z}_j - \mathbf{x}_i^m(t))}{\sum_{j=1}^{n_p} k(\mathbf{z}_j - \mathbf{x}_i^m(t)) w_j}, \quad (17)$$

where $k(\cdot)$ is the kernel function, $n_p$ is the number of pixels in the part and $w_j$ is the sample weight at $\mathbf{z}_j$:

$$w_j = -\frac{\partial}{\partial \mathbf{x}_i^m}[d(\mathbf{P}_i, \phi(I_i^m)) + \sum_{j \neq i} v_{i,j}^m]\bigg|_{\mathbf{x}_i^m = \mathbf{z}_a}. \quad (18)$$

The iteration stops when the magnitude of mean-shift vector $\|\mathbf{x}_i^m(t+1) - \mathbf{x}_i^m(t)\|$ is small enough.

*2) Part Size Fitting*

The goal is to optimize the part size while fixing the appearance $\mathbf{P}$ and location $\mathbf{X}$. Same as the part localization step, the discrimination cost term from (7) is held constant since $\mathbf{P}$ is fixed. Problem (1)–(3) can thus be written as

$$\min_{\mathbf{S}} \sum_{m=1}^{N}\left(\sum_{i=1}^{K} d(\mathbf{P}_i, \phi(I_i^m)) + \sum_{i=1}^{K}\sum_{j=1}^{K} v_{i,j}^m\right) \quad (19)$$

$$\text{s.t. } 0 \leq \mathbf{s}_i \leq 1, \ i = 1, ..., K \quad (20)$$

$$0 \leq \mathbf{x}_i \pm (1/2)\mathbf{s}_i \leq 1, \ i = 1, ..., K. \quad (21)$$

Here we solve for $\mathbf{S}$ by adopting the scale-space mean-shift algorithm [31]. In the formulation, given the coordinate base $b > 1$, the part size is iteratively updated by

$$\mathbf{s}_i^m(t+1) = \mathbf{s}_i^m(t) b^{r'}, \ r' = \frac{\sum_{r \in \Omega} \sum_{j=1}^{n_p} H(\mathbf{z}_j, r) w(\mathbf{z}_j) r}{\sum_{r \in \Omega} \sum_{j=1}^{n_p} H(\mathbf{z}_j, r) w(\mathbf{z}_j)}, \quad (22)$$

where $\Omega$ is the search range in the scale space centered at the current part size $\mathbf{s}_i^m(t)$, $H$ is the scale kernel, $n_p$ is the number of pixels inside the current part size, and $w(\mathbf{z}_a)$ is the sample weight defined in (18). The iteration stops when $|r'|$ is small enough.

*3) Part Model Learning*

The goal here is to find the optimal part appearance $\mathbf{P}$ without changing location $\mathbf{X}$ and size $\mathbf{S}$. When optimizing (8), the separation cost term from (5) is constant with fixed $\mathbf{X}$ and $\mathbf{S}$. Therefore the part appearance model $\mathbf{P}_i$ can be found by

$$\min_{\mathbf{P}_i} \sum_{m=1}^{N} d(\mathbf{P}_i, \phi(I_i^m)) - \sum_{j=1}^{K} d(\mathbf{P}_i, \mathbf{P}_j). \quad (23)$$

Note that if the discrimination cost term from (7) were ignored, the objective function would be $\sum_{m=1}^{N} d(\mathbf{P}_i, \phi(I_i^m))$, which only considers the similarity between the model and all image regions. The solution for $\mathbf{P}_i$ would then be given by the average of $\phi(I_i^m), m = 1, ..., N$ when $L^2$-distance is used.

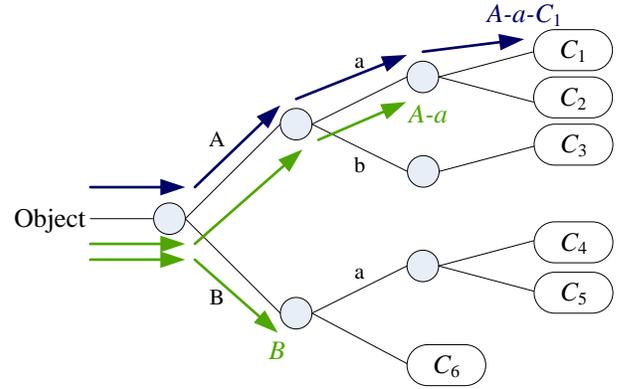

Fig. 3. An example of hierarchical partial classifier. The class hierarchy is learned from the training data via an unsupervised clustering procedure. The fully-classified instance (blue) reaches the leaf layer and receives a complete label sequence *A-a-$C_1$*, while the ambiguous instances (green) stop at middle layers and receive incomplete sequences *A-a* and *B*.

## V. HIERARCHICAL PARTIAL CLASSIFIER

To exploit the information from uncertain data without introducing misclassification, we develop a novel technique that learns a hierarchical structure for the classes and allows for indecision for ambiguous data. A class hierarchy, i.e., a binary decision tree with one classifier at each node, is generated to determine the grouping of classes in higher levels. The grouping labels can serve as coarse categorization results when the exact class label cannot be identified. In the testing phase, the input data instance is examined by layers of classifiers, each of which gives a prediction label. If the instance falls in the indecision range at any layer, the classification procedure stops and returns an incomplete sequence of class labels. In this way, misclassifications are avoided without losing the entire information provided by uncertain data. The concept of hierarchical partial classifier is illustrated in Fig. 3.

*A. Unsupervised Construction of Class Hierarchy*

The class hierarchy follows a binary tree structure, i.e., each node separates data into two categories. The arrangement of class grouping is learned by an unsupervised recursive clustering procedure as follows. The EM algorithm for mixture of Gaussians (MoG) is applied to separate all data into two clusters, which can be viewed as "positive" and "negative" data respectively. For each species, data are relabeled based on which cluster the majority of this species belongs. A radial bases function (RBF) kernel support vector machine (SVM) is trained with these two super-classes. The above steps are then repeated separately within each cluster until there is only one species in each cluster.

To handle the class imbalance issue, which is caused by the dominance of one or more species in the sampled habitats, a biased-penalty approach is adopted during the SVM training procedure [32]. Rather than using only a single penalty parameter, the penalty parameters for positive and negative classes are set differently to $C_+ = C \cdot N_- / N_{total}$ and $C_- = C \cdot N_+ / N_{total}$, where $C$ is the original penalty parameter, $N_+$ and $N_-$ denote separately the number of positive and negative training samples, and $N_{total} = N_+ + N_-$.

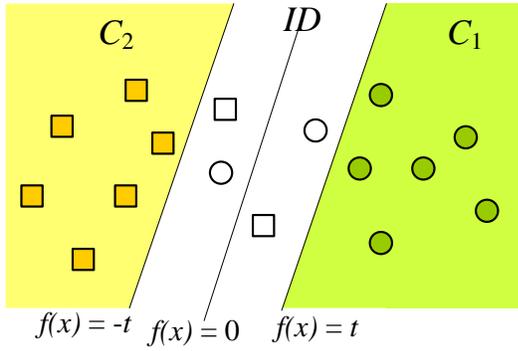

Fig. 4. Partial classification for an SVM. Ambiguous data have small absolute decision values, so they fall in the indecision domain (ID) and are not assigned to either of the classes. This figure is best viewed in color.

### B. Benefit-based Partial Classification

After the SVM classifier is trained, one needs to define its indecision criterion in order to enable partial classification. In light of evaluating deferred decisions, our task is formulated as an optimization problem as follows. Given the data $(\mathbf{x}_i, y_i)$, $i = 1,...,N$, and an SVM decision function $f: \mathbf{R}^d \to \mathbf{R}$ trained by these data, the generalized benefit function of partial classification is defined as

$$B(D) = s_c(\mathbf{x}) P(D, \hat{y} = y) - s_w(\mathbf{x}) P(D, \hat{y} \neq y), \quad (24)$$

where $s_c(\mathbf{x})$ and $s_w(\mathbf{x})$ are score functions for correct and wrong decisions, respectively, and $D$ denotes the event of decisions being made. One can interpret (24) as the expected value of total reward for classification, where one earns $s_c(\mathbf{x}_i)$ points for being correct, loses $s_w(\mathbf{x}_i)$ points for being wrong, and gets zero points for indecision with the $i$-th data point. The score functions can be any nonnegative functions that decrease monotonically with respect to $|f(\mathbf{x})|$ so that a greater importance is added to ambiguous data. We hence choose $s_c(\mathbf{x}) = s_w(\mathbf{x}) = \exp(-|f(\mathbf{x})|)$.

The goal is to find $D$ that maximizes (24). First note that $y \in \{-1, 1\}$. Also, a correct decision implies $yf(\mathbf{x}) > 0$ and a wrong decision implies $yf(\mathbf{x}) < 0$. As illustrated in Fig. 4, a partial SVM classifier makes a decision only if $|f(\mathbf{x})|$ is greater than a threshold $t > 0$. Therefore (24) can be written as

$$B(t) = e^{-yf(\mathbf{x})} P(yf(\mathbf{x}) \geq t) - e^{yf(\mathbf{x})} P(yf(\mathbf{x}) \leq -t)$$
$$= \frac{1}{N} \left( \sum_{i=1}^{N} e^{-a_i} \mathbf{1}[t \leq a_i] - \sum_{i=1}^{N} e^{a_i} \mathbf{1}[t \leq -a_i] \right), \quad (25)$$

where $a_i := y_i f(\mathbf{x}_i)$ and $\mathbf{1}[\cdot]$ denotes the 0-1 indicator function. It can be easily verified, as shown in Fig. 5, that indicator

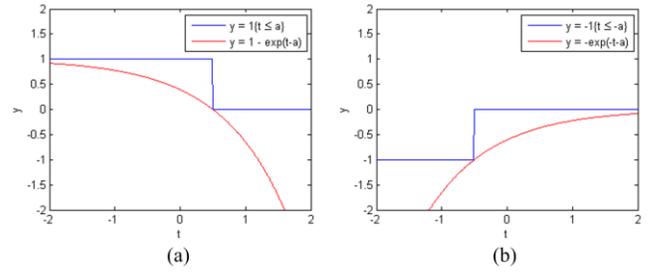

Fig. 5. Visualization of (a) Equation (26) and (b) Equation (27) with $a = 0.5$.

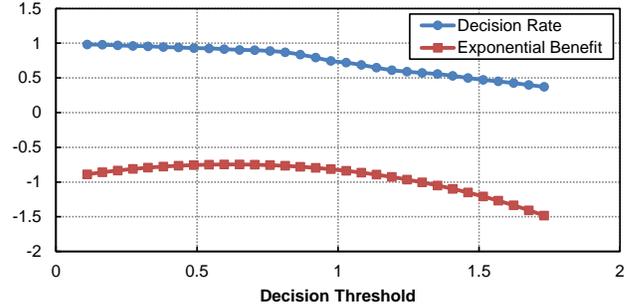

Fig. 6. Decision rate and exponential benefit vs. decision threshold.

functions are bounded below by exponential functions, i.e.,

$$\sum_{i=1}^{N} \mathbf{1}[t \leq a_i] \geq \sum_{i=1}^{N} (1 - e^{t-a_i}), \quad (26)$$

$$-\sum_{i=1}^{N} \mathbf{1}[t \leq -a_i] \geq -\sum_{i=1}^{N} e^{-t-a_i}. \quad (27)$$

Using (26) and (27), we define an exponential benefit function, $B_{exp}(t)$, which serves as a lower bound of $B(t)$:

$$B_{exp}(t) = (1/N)(\sum_{i=1}^{N} e^{-a_i}(1 - e^{t-a_i}) - \sum_{i=1}^{N} e^{-a_i} e^{-t-a_i})$$
$$= (1/N)(\sum_{i=1}^{N} e^{-a_i} - e^t \sum_{i=1}^{N} e^{-2a_i}) - e^{-t}. \quad (28)$$

An example of the exponential benefit function with respect to decision threshold is shown in Fig. 6. Based on this, selecting the decision threshold can be written as an inequality constrained minimization problem:

$$\min_t \quad e^t \sum_{i=1}^{N} e^{-2a_i} + N e^{-t} \quad (29)$$

$$\text{s.t.} \quad f_{min} \leq t \leq f_{max}, \quad B_{exp}(t) \geq B_{exp}(0), \quad (30)$$

where $f_{min} = \min_{i=1,...,N} |f(\mathbf{x}_i)|$ and $f_{max} = \max_{i=1,...,N} |f(\mathbf{x}_i)|$. Constraints in (30) ensure not only feasible solutions but also a gain in the exponential benefit function comparing to full classification. The problem defined in (29), (30) is solved by applying the barrier method [33]. The optimal threshold $t^*$ found by solving (29), (30) is then used in the testing phase.

## VI. EXPERIMENTAL RESULTS

### A. Datasets

The proposed method is evaluated on the Fish4Knowledge

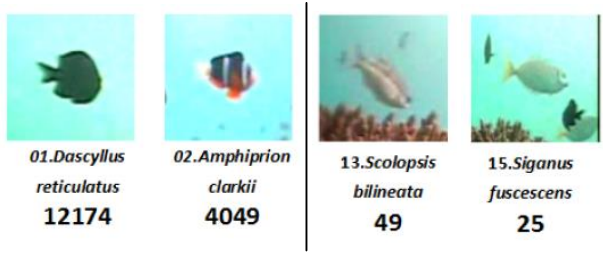

Fig. 7. Two largest species (left) and two smallest species (right) from Fish4Knowledge dataset.

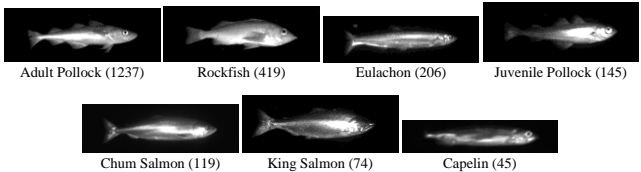

Fig. 8. Fish species in NOAA Fisheries dataset sorted in the descending order of the number of samples.

(F4K) recognition dataset [34]. For comparison purpose, we followed the settings in [4] and conducted the experiments on the top 15 species, which consists of 26418 fish images. As shown in Fig. 7, the dataset is highly imbalanced where the data size of the most frequent species is approximately 500 times of the least frequent species. Object segmentation mask for each fish image is provided along with the dataset. We would like to emphasize again that our feature learning method remains fully unsupervised even by adopting segmentation, and existing unsupervised techniques such as the GrabCut [27] can be applied to generate segmentation masks on the fly.

Extending our previous work on video-based fisheries surveys [10], we also evaluate the proposed method on NOAA Fisheries dataset, a self-collected underwater fish dataset as shown in Fig. 8. The images are captured by the Cam-trawl system [7] from a mid-water trawl with only simple web patterns in the background and being illuminated by artificial LED lighting. The dataset consists of 2195 grayscale fish images from 7 species. All fish images are manually labeled by following instructions from fisheries scientists. Compared to Fish4Knowledge dataset, NOAA Fisheries dataset collects images at higher resolution, but discard color information due to the color distortion in deeper water. We apply our automatic fish segmentation algorithm [8] to generate reliable object bounding boxes and segmentation masks. In order to persist the generality, the proposed species recognition framework does not use stereo vision despite the Cam-trawl is a stereo camera system.

### B. Implementation Details

Before the analysis, each image is scaled so that the bounding box is no larger than $200 \times 200$ pixels with its aspect ratio preserved. The number of parts is empirically determined as $K = 6$ in the experiments (more discussion in Section VI.E). Each part is initialized with a size of $48 \times 48$ pixels in the rescaled images. For part features $\mathbf{P}$, the SIFT descriptors and weighted color histogram is used. SIFT descriptors are sampled densely every 4 pixels within the part region. After extraction, the dimensionality is reduced to 128 by applying the principal component analysis (PCA). The weighted histogram is the HSV color histogram weighted by an isotropic kernel that is maximized at the center [35]. Geometric information of parts is also encoded by the location and size of each part in the normalized coordinates $\mathbf{x}_i^m, \mathbf{s}_i^m \in [0,1] \times [0,1]$. In addition to localized features, global features are also taken into account during classifier training. The same SIFT descriptors and weighted color histogram are extracted from the entire bounding box. The global features and local features from each part are concatenated to form the feature vector for one image.

TABLE I
ACCURACY OF FEATURE LEARNING METHODS ON NOAA FISHERIES

| Method | Accuracy (%) |
|---|---|
| Template [14] | 88.4 |
| Alignment [15] | 91.6 |
| PAFF [25] | 90.1 |
| Proposed (grid, spatial-order) | 86.9 |
| Proposed (saliency, spatial-order) | 89.5 |
| Proposed (saliency, relax-label) | **93.8** |
| Proposed (removing pectoral fins) | 87.3 |

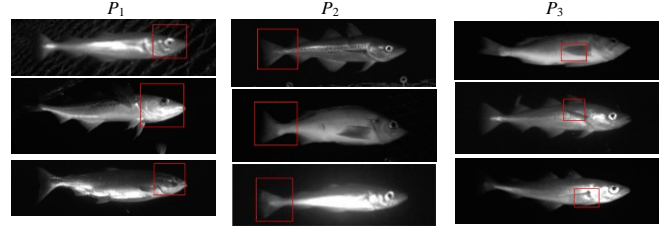

Fig. 9. Examples of parts detected by the proposed unsupervised feature learning algorithm. Each column shows the parts found by one part model. Meaningful fish body parts are successfully learned regardless changes in size, such as head, caudal fin and pectoral fin.

Parameters in the unsupervised non-rigid part model learning algorithm are set empirically as follows. For the stopping criteria in Algorithm 1, we set the convergence of object part models as $\sum_{i=1}^{K} \|\Delta \mathbf{P}_i\| \leq 0.5$. The maximum iteration is set to 15 as we observed that the algorithm converges within 10 iterations in most cases. The part initialization algorithm is not very dependent on the Gaussian standard deviation $\sigma$. A typical value such as 0.5 or 1 (i.e., a $3 \times 3$ or $5 \times 5$ filter window) works in most cases. The convergence criteria for mean-shift update in part localization is set to $\varepsilon_\mathbf{x} = 0.5$. As for part size fitting, the coordinate base is set to $b = \sqrt[5]{2}$ and the search space as $-2 \leq \omega \leq 2$. The choices for parameters in our implementation depend highly on the characteristics of images, correlation between training and testing samples, as well as the computational consumption. To achieve better accuracy in feature descriptors, an object part is usually initialized with a square with each side at least 0.2 of the longest side of bounding box. Also the convergence criterion for training the non-rigid part model should be no more than 0.1 of the average norm of part feature vectors times the number of parts. In our experiments the parameters for part size fitting follows the work on scale-space mean-shift algorithm [31].

The method for solving (23) in part model learning step depends on the distance metric chosen for features. In our experiments, we measure the distance metric between feature descriptors $\mathbf{P}$ and $\mathbf{Q}$ by the normalized correlation function, i.e., $d(\mathbf{P},\mathbf{Q}) = 1 - \mathbf{p}^T\mathbf{q}$, where $\mathbf{p} = \mathbf{P}/\|\mathbf{P}\|$ and $\mathbf{q} = \mathbf{Q}/\|\mathbf{Q}\|$. Hence (23) can be minimized by solving a standard linear programming problem:

TABLE III
PERFORMANCE ON NOAA FISHERIES DATASET

| Method | AP (%) | AR (%) | AC (%) |
|---|---|---|---|
| Flat SVM | 87.9 | 83.1 | 86.6 |
| Proposed | **97.1** | **98.9** | **98.4** |

$$\min_{\mathbf{p}_i, c} \sum_{j=1}^{K} \mathbf{p}_i^T \mathbf{p}_j \quad (31)$$

$$\text{s.t.} \quad \mathbf{p}_i^T \mathbf{i}_i^m \geq c, m = 1,...,N, \quad (32)$$

where $\mathbf{p}_i = \mathbf{P}_i/\|\mathbf{P}_i\|$ and $\mathbf{i}_i^m = \phi(I_i^m)/\|\phi(I_i^m)\|$.

We use the LIBSVM [36] to train the multi-class classifier, which is implemented by the one-versus-one strategy. The 10-fold cross-validation is applied, and the final classifier is used for the testing set.

*C. Feature Learning*

In this experiment, the proposed non-rigid part model learning algorithm is compared with several feature descriptors or learning methods. The template model [14] and alignment method [15] are unsupervised techniques, while the part-aware fish features (PAFF) [25] is a supervised technique. Fish images are classified by a flat multiclass SVM, which simultaneously classifies 7 species from the NOAA Fisheries dataset.

The accuracy rates is shown in Table I. The proposed method, which is based on saliency and relaxation labeling, outperforms the fine-grained object recognition methods as well as the supervised PAFF method. Also, the methods based on simple grid-based part initialization and spatial ordering for part matching are compared in Table I. For grid initialization we used an 8-part model as in [22], which consists of one block for the head, one block for the tail and a $3 \times 2$ grid for the torso. Spatial ordering matches the parts purely based on the part locations. These two approaches are followed by the proposed unsupervised part model learning algorithm proposed in this paper. As shown in Table I, these approaches give lower accuracy rates than the proposed method for not taking into account the potential rotation or deformation of fish body, which are common in images captured from unconstrained natural habitats.

A visualization of some fish parts discovered by the proposed algorithm are shown in Fig. 9. In addition to the head and the caudal fin (tail), which are most seminal fish parts, the non-rigid part model locates the pectoral fin and classify the fish based on its characteristics. In particular, the pectoral fin part is systematically discovered by the unsupervised non-rigid part model learning. If we remove the corresponding features deliberately, the recognition performance is decreased, as shown in the last row of Table I. This serves as a justification for the high impact of salient features systematically discovered by the proposed non-rigid part model.

TABLE II
PERFORMANCE ON TOP 15 SPECIES OF FISH4KNOWLEDGE DATASET

| Method | AP (%) | AR (%) | AC (%) |
|---|---|---|---|
| Flat SVM | 88.5 | 76.9 | 95.7 |
| PCA-Flat SVM | 88.9 | 77.7 | 95.4 |
| CART [37] | 52.9 | 53.6 | 87.0 |
| Taxonomy | 87.2 | 76.1 | 95.3 |
| BEOTR [4] | 91.4 | 84.8 | 97.5 |
| Proposed | **92.1** | **91.6** | **97.7** |

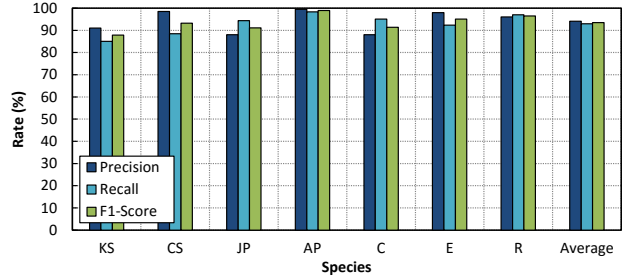

Fig. 10. Precision, recall and $F_1$-score of each species from NOAA Fisheries.

*D. Hierarchical Partial Classification*

In this experiment, the proposed hierarchical partial classifier algorithm is evaluated with both NOAA Fisheries and Fish4Knowledge datasets. The proposed algorithm is compared with several baseline algorithms including Flat SVM, principal component analysis (PCA-Flat SVM), taxonomy tree, BEOTR [4] and the classification and regression tree (CART) [37]. A 10-fold cross-validation procedure is applied to train the classifiers. The trajectory voting scheme from [4] is applied to the prediction labels so that results in the same trajectory remain consistent.

Results of Fish4Knowledge and NOAA Fisheries dataset are shown respectively in Table II and Table III. The proposed hierarchical partial classifier performs the best in terms of precision, recall as well as accuracy. The recognition performance of each algorithm is evaluated through the average precision (AP), average recall (AR) in addition to the accuracy (AC). The metrics AP and AR are defined as follows:

$$AP = \frac{1}{C}\sum_{i=1}^{C} precision(i) = \frac{1}{C}\sum_{i=1}^{C}\frac{TP_i}{TP_i + FP_i} \quad (33)$$

$$AR = \frac{1}{C}\sum_{i=1}^{C} recall(i) = \frac{1}{C}\sum_{i=1}^{C}\frac{TP_i}{TP_i + FN_i} \quad (34)$$

where $TP_i$, $FP_i$ and $FN_i$ denote the $i$-th class true positive, false positive and false negative, respectively. Same as in [25], only the complete classifications are computed in $TP_i$, $FP_i$ and $FN_i$ for every tested methods. Precision reflects the percentage of correct data in a certain recognized class. Recall indicates the percentage of data from a certain class being correctly recognized. In addition, the precision, recall and $F_1$-score for each species is reported in Fig. 10. The $F_1$-score, which provides an integrated measure combining the precision and

recall, is given by

$$F_1\text{-}score = \frac{2 \cdot precision \cdot recall}{precision + recall}. \quad (35)$$

The proposed algorithm achieves an average $F_1$-score of 93.4% and demonstrates consistency among fish species, as shown in Fig. 10. The partial decision (PD) rate for the proposed method is 4.92% (more discussion in Section VI.E), which is acceptable for most scenarios of fish species identification.

*E. Discussion*

The proposed non-rigid part learning and partial classification framework builds up an error-resilient object recognition system that is applicable "in the field," where most of the images are noisy and with substantial degree of ambiguity. Low contrast in underwater imagery, for example, weakens the edges and texture and thus introduces error to part appearance. Pose variation gives a different view of the part and results in deviation when describing part size and appearance. Features with such error or deviation can be regarded as "outliers" from its species and are likely to fall on the opposite side of the SVM decision boundary, as shown by the white instances in Fig. 4. The proposed hierarchical partial classification reduces misclassification by avoid making guesses with low confidence and thus enhances the recognition performance in practical datasets. Moreover, the extent of conservativeness of the proposed classifier is highly adaptive since the indecision region is optimized based on the distribution of data. This makes the proposed classifier intelligent and fully automatic that requires no manual interference by the user.

In many cases, the performance of unsupervised learning algorithms depends highly on how well the variables are initialized. For the proposed non-rigid part model, one can decide the number of parts to be learned by the part model. This factor not only affects the power of discrimination but also gives different dimensionality of feature descriptors that represent fish species characteristics.

To investigate into this, the proposed algorithm is tested with different numbers of parts. As shown in Table IV, the best performance is achieved by using 6 parts. It successfully learns rather subtle but meaningful parts of fish body such as the pectoral fins, which are consistent with the domain knowledge in fisheries studies for recognizing a fish species. Also noted that the performance is decreased by using 8-part and 10-part model. The reason is that these models tend to divide some large seminal parts (e.g., head and tail) into small pieces, which leads to worse performance when locating parts. Moreover, the models with more parts are likely to incur overlapping of parts and thus the discrimination cost is greatly increased.

As for classification, the proposed hierarchical partial classifier is able to handle uncertain or missing data, which is a common and challenging issue in practical applications of object recognition. For example, capturing images for freely-swimming fish in an unconstrained environment usually introduces a high uncertainty in many of the data due to poor capture quality and non-lateral fish. To see the effectiveness of handling uncertain data by partial classification, we compare the performance using a flat classifier, a hierarchical full classifier and the proposed method. The flat classifier is a multi-class one-against-all SVM, which classifies objects to all classes simultaneously. The hierarchical full classifier follows the proposed method from learning the class hierarchy to training every SVM, except the decision threshold is set to zero so that all instances are classified down to the lowest layer. The accuracy and partial decision rate (percentage of data not being classified to the lowest level) of the proposed algorithm are shown in Table V. The flat classifier performs the worst due to the misclassification for those images with high uncertainty. Besides, the subtle visual difference between some species is difficult to learn by a single classifier. The hierarchical classifier learns the relations directly from the data, and thus performs better when the inter-class variation is small. By allowing partial classification using the optimal decision criterion, the accuracy is further increased by 4% while less than 5% of data receive incomplete categorizations.

TABLE IV
PERFORMANCE VS. NUMBER OF PARTS

| Model | AP (%) | AR (%) | AC (%) |
|---|---|---|---|
| 4 parts | 87.9 | 83.1 | 86.6 |
| 6 parts | **97.1** | **98.9** | **98.4** |
| 8 parts | 91.7 | 90.8 | 92.8 |
| 10 parts | 74.6 | 72.9 | 78.3 |

TABLE V
FULL VS. PARTIAL CLASSIFICATION ALGORITHMS

| Classification | AC (%) | PD (%) |
|---|---|---|
| Flat SVM | 93.8 | - |
| Hierarchical Full SVM | 94.3 | 0.00 |
| Hierarchical Partial SVM | 98.4 | 4.92 |

VII. CONCLUSION

In this paper, a novel framework for underwater fish recognition is proposed. The proposed framework is facilitated by unsupervised learning algorithms and thus reduces the requirement of human interference comparing to existing approaches. The non-rigid part model effectively discovers discriminative parts by adopting saliency and relaxation labeling. Fitness, separation and discrimination of parts are considered for finding meaningful representations of fish body parts in a fully unsupervised fashion. On the other hand, data uncertainty and class imbalance are two of the most common issues in practical classification applications. The proposed hierarchical partial classification successfully handled these issues by enabling coarse-to-find categorization and thus retrieving partial information from those ambiguous data which are possibly misclassified or rejected by other algorithms. We further develop a systematic optimization approach to selecting decision criteria for partial classifiers by introducing the exponential benefit function. Experimental results show a favorable performance of fish recognition on both large-scale public dataset and practical highly-uncertain dataset of live fish. Future work includes investigation of how to make stereo imaging and object part matching helpful to each other and conducting more tests of the proposed algorithm with different types of objects.